\documentclass[journal]{IEEEtran}

\usepackage[utf8]{inputenc}
\usepackage[T1]{fontenc}
\usepackage{amsmath,amsfonts}
\usepackage{tabularx}
\usepackage{algorithmic}
\usepackage{algorithm}
\usepackage{array}
\usepackage[caption=false,font=normalsize,labelfont=sf,textfont=sf]{subfig}
\usepackage{textcomp}
\usepackage{stfloats}
\usepackage{url}
\makeatletter
\g@addto@macro{\UrlBreaks}{\UrlOrds}
\makeatother
\usepackage{verbatim}
\usepackage{graphicx}
\usepackage{tikz}
\usetikzlibrary{arrows.meta, positioning, fit, calc, backgrounds}
\usepackage{float}
\usepackage{cite}
\usepackage{hyperref}
\usepackage{booktabs}               
\usepackage[table]{xcolor}
\usepackage{threeparttable}
\usepackage{multirow}               
\usepackage{makecell}               
\usepackage{pifont}                 
\usepackage{xcolor}

\newcommand{\expminus}{\rule[0.5ex]{0.30em}{0.5pt}\mkern1mu}

\newif\ifuseRevisions
\useRevisionsfalse  

\usepackage{xcolor}
\usepackage{graphicx}
\usepackage[most]{tcolorbox}

\newcommand{\rev}[1]{\ifuseRevisions\textcolor{blue}{#1}\else #1\fi}

\begin{document}

\title{Accurate Trajectory Tracking with \rev{Model Predictive Contouring Control} for \rev{Bird-Scale} Flapping-Wing MAVs}

 \author{Charbel Toumieh, Jack Zeng, Niel Mistry, Dario Floreano,~\IEEEmembership{Fellow,~IEEE}
\thanks{Accepted for publication in \emph{IEEE Robotics and Automation Letters}.}%
\thanks{\textcopyright~2026~IEEE\@. Personal use of this material is permitted. Permission from IEEE must be obtained for all other uses, in any current or future media, including reprinting/republishing this material for advertising or promotional purposes, creating new collective works, for resale or redistribution to servers or lists, or reuse of any copyrighted component of this work in other works.}%
\thanks{The authors are with the Laboratory of Intelligent Systems, Ecole Polytechnique Federale de Lausanne (EPFL), CH1015 Lausanne, Switzerland.}%
 \thanks{This work was supported by the Swiss National Science Foundation (SNSF) with grant number 200020\textunderscore212078, and Armasuisse grant number 591797.}
 \thanks{Video: \url{https://youtu.be/pVb3DoUntWI}}
 \thanks{Code: \url{https://github.com/lis-epfl/xfly_stack}}
 }%

\maketitle
\thispagestyle{empty}
\pagestyle{empty}

\begin{abstract}
Flapping-wing micro aerial vehicles offer quieter and safer operation than rotary-wing drones, yet achieving precise autonomous control of bird-scale ornithopters remains challenging: lift, airspeed, and turning authority are tightly coupled and governed by only a few control inputs. Conventional cascaded controllers treat altitude, speed, and heading independently, producing persistent tracking errors during complex maneuvers, while time-parameterized trajectory tracking requires predefined speed profiles that existing methods cannot robustly produce for these coupled dynamics. We address both limitations simultaneously with a Model Predictive Contouring Control (MPCC) approach that tracks arc-length-parameterized trajectories while optimizing progress online, eliminating the need for predefined timing. However, MPCC requires a dynamical model that captures the coupled aerodynamics without exceeding the computational budget of real-time nonlinear optimization. Here, we propose a compact, continuously differentiable model that captures the dominant couplings of bird-scale ornithopters, enabling real-time predictive control. We validated the method with the XFly ornithopter flying along circular and three-dimensional racing trajectories and achieved a mean deviation from the reference trajectory between 6.5 and 9\,cm at speeds up to 3\,m/s, which represents an \rev{8.5$\times$} improvement over prior ornithopter control methods.
\end{abstract}


\section{INTRODUCTION}

In contrast to conventional \rev{propeller-driven drones, from rotorcraft to avian-inspired
morphing-wing platforms~\cite{phan2024twist,jeger2024adaptive,wanner2026flight}}, flapping-wing micro aerial vehicles offer quieter operation and safer physical interaction due to the absence of exposed propellers \cite{floreano2015science}. Remarkable control autonomy has been achieved at the insect scale \rev{\cite{ma2013controlled, jafferis2019untethered, chen2021collision, hsiao2025aerobatic, hsiao2026onboard, bena2023beepp}}, but controlling bird-scale ornithopters presents a fundamentally different challenge. Unlike insect-scale ornithopters that can generate lift independently of forward motion \rev{\cite{wu2025hover}}, bird-scale ornithopters generate lift during forward flight and are affected by airspeed. Recent work has advanced their mechanical design and onboard sensing \cite{chin2020efficient, zufferey2021design, kim2024wingstrain, rafee2025review}, yet precise trajectory control remains an open problem.

\begin{figure}
\centering
  \includegraphics[width=\columnwidth, trim={0cm 0cm 0cm 0cm}, clip]{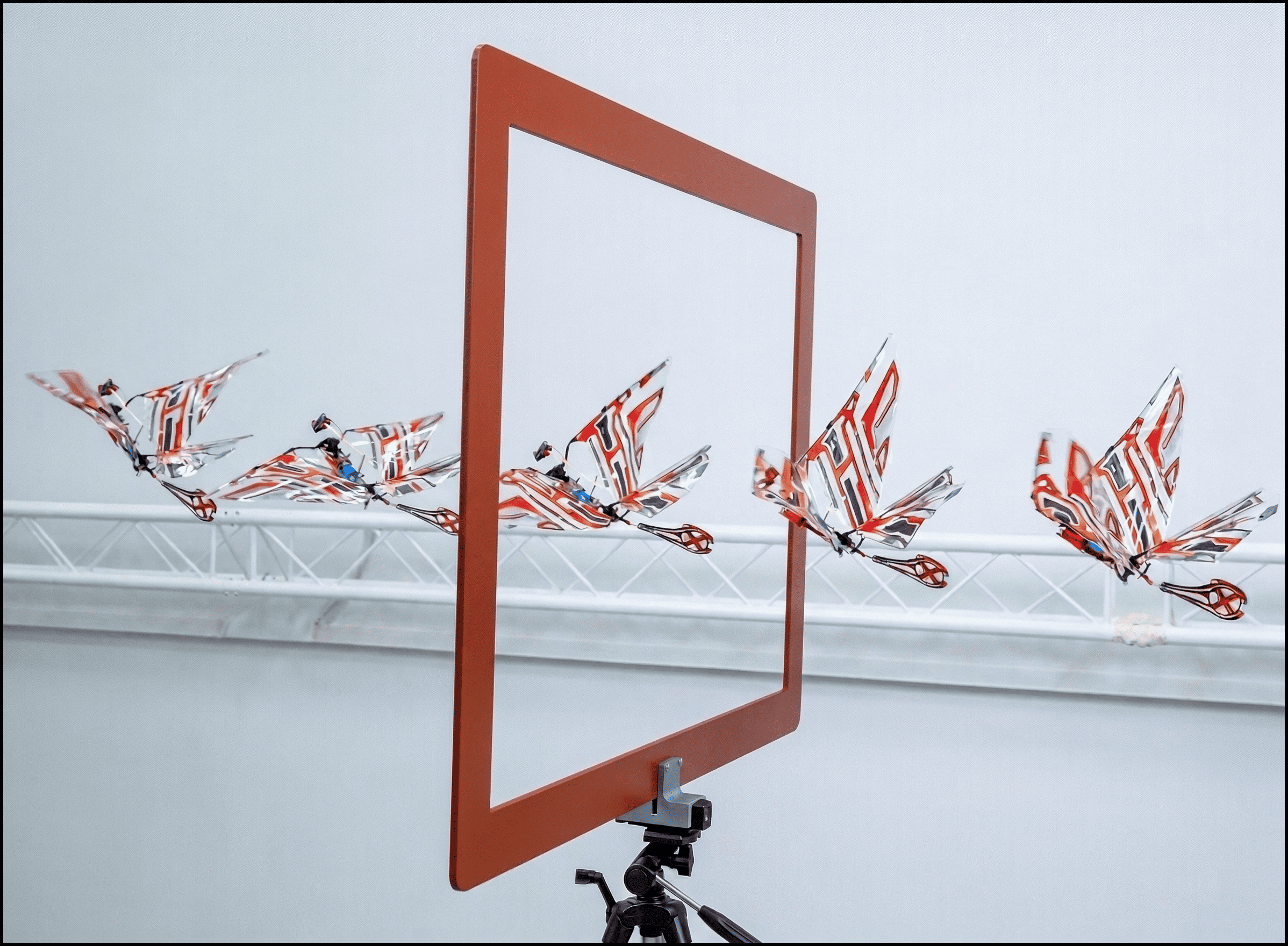}
  \caption{
    Time-lapse of the ornithopter passing through a gate during trajectory tracking. 
  }
  \label{fig:drone_gate}
\end{figure}
Governed by a small number of control inputs---typically flap intensity and rudder deflection---these vehicles exhibit strong coupling: increasing flapping effort simultaneously accelerates the vehicle and induces climb, while increasing airspeed enhances turning authority. Existing flight control methods do not adequately address this intrinsic coupling. At the control level, cascaded PID loops process altitude, speed, and heading independently \cite{ndoye2023vector,  huang2022ustbird,escobar2025dynamical,   jiang2026sparrowhawk}. These reactive loops cannot anticipate how correcting one variable (e.g., turning)  perturbs another (e.g., inducing altitude loss) and therefore yield persistent tracking errors during complex maneuvers. A related challenge arises at the planning level. Most controllers treat flight as time-parameterized trajectory tracking \cite{mellinger2011minimum}, which assumes that a feasible speed profile is known in advance. For ornithopters, feasible speed is dictated by coupled climb and turn-radius constraints, making pre-specified time-parameterized trajectories fragile: any deviation from the planned speed — caused by battery depletion, aerodynamic disturbances, or modeling error — produces large along-track errors that cascade into altitude and heading tracking through the couplings. Model Predictive Contouring Control (MPCC) \cite{liniger2015mpcc} addresses both issues: it removes the need for pre-specified timing by tracking an arc-length-parameterized trajectory while optimizing progress online, and it resolves the couplings in real time by re-solving the control problem at every cycle.

However, the application of predictive control requires a dynamical model that captures the coupled dynamics without exceeding the computational limits of real-time optimization. Unsteady CFD and vortex-particle models that resolve the flapping wake run orders of magnitude slower than real-time \cite{shyy2013introduction}, while simplified models neglect the couplings that the predictive framework is designed to handle \cite{escobar2025dynamical, ndoye2023vector}. Here we bridge this gap with a nine-state dynamical model that captures the dominant aerodynamic couplings of ornithopter flight while using only continuously differentiable functions---a requirement for gradient-based nonlinear programming solvers that make real-time MPCC tractable---and keeping the number of decision variables per time step low enough to solve within a 10\,ms control period. We embed this model within an MPCC formulation, creating a predictive framework that directly addresses the platform's coupled aerodynamic constraints without requiring a pre-specified time-parameterized speed profile. Finally, we experimentally validate the control method on the XFly ornithopter (by Bionic Bird\rev{; Fig. \ref{fig:drone_gate}}), a commercially available drone with flapping wings and a tail equipped with a vertical rudder. We show precise, real-time trajectory following on circular and three-dimensional racing trajectories that achieve a mean deviation
from the reference trajectory (cross-track error) between 6.5 and 9\,cm (8.5$\times$ smaller than the state of the art \cite{escobar2025dynamical}) at speeds up to 3\,m/s.

\rev{Specifically, the contributions of this letter are:
    (i)~a compact, continuously differentiable nine-state model that captures
    the dominant aerodynamic couplings of bird-scale ornithopter
    flight while remaining low-dimensional
    enough for real-time nonlinear optimization;
    (ii)~to our knowledge the first model predictive contouring control
    formulation for bird-scale flapping-wing flight, which tracks an
    arc-length-parameterized trajectory and thereby eliminates the need for a
    pre-specified speed profile while resolving the platform's couplings online;
    (iii)~experimental validation on the XFly ornithopter, demonstrating an
    8.5$\times$ reduction in cross-track error over the prior state of the
    art~\cite{escobar2025dynamical} on circular trajectories and accurate
    tracking of three-dimensional racing trajectories with altitude variation, a
    trajectory class not previously demonstrated on this platform; and
    (iv)~an ablation study that quantifies the contribution of each modeling
    component to closed-loop tracking accuracy.}
\section{Methods}
\rev{The overall control architecture is summarized in Fig.~\ref{fig:control_arch}. More details on each component are provided in the subsections below.}


\newsavebox{\controlarchbox}
\savebox{\controlarchbox}{%
\begin{tikzpicture}[
    >=Stealth,
    every node/.style={font=\footnotesize},
    B/.style={
        draw=#1!55!black, rounded corners=1pt, line width=0.5pt,
        align=center, fill=#1!8, inner xsep=4pt, inner ysep=2pt,
        minimum height=8mm, font=\footnotesize
    },
    B/.default=black,
    sig/.style={->, line width=0.6pt},
    lbl/.style={font=\footnotesize, inner sep=1pt},
    G/.style={draw=blue!30!black, rounded corners=2pt,
              inner xsep=4pt, inner ysep=3pt, line width=0.4pt, fill=blue!3},
]

\node[B=blue] (traj) {\textbf{Traj.\ generation}\\Quintic B\'{e}zier};

\node[B=blue, below=5.5mm of traj, minimum height=11.5mm] (mpcc) {\textbf{MPCC}\\9-state model\\Knitro\;\textperiodcentered\;100\,Hz};

\node[B=blue, below=5.5mm of mpcc] (est) {\textbf{State estimation}\\Kalman filtering};

\node[B=green!60!black] (wmap) at (traj-|{$(traj.east)+(36mm,0)$}) {\textbf{World map}\\Gates, obstacles};

\node[B=orange] (plant) at (mpcc-|wmap) {\textbf{XFly ornithopter}\\Steer/Straight assist};

\node[B=gray] (mocap) at (est-|wmap) {\textbf{Motion capture}\\OptiTrack\;\textperiodcentered\;240\,Hz};

\begin{scope}[on background layer]
\node[G, fit=(traj)(mpcc)(est)] {};
\end{scope}

\draw[sig] (wmap.west)--(traj.east) node[lbl,above,midway]{constraints};

\draw[sig] (traj.south)--(mpcc.north) node[lbl,right,midway,xshift=0.75mm]{$\mathbf{r}(\theta)$};

\draw[sig] (est.north)--(mpcc.south) node[lbl,right,midway,xshift=0.75mm]{$\hat{\mathbf{x}},\;\hat{\theta}$};

\draw[sig] ([yshift=1.5mm]mpcc.east)--([yshift=1.5mm]plant.west) node[lbl,above,midway]{$u_{\mathrm{flap}},\;u_{\mathrm{rud}}$};

\draw[sig,orange!55!black] ([yshift=-1.5mm]plant.west)--([yshift=-1.5mm]mpcc.east) node[lbl,below,midway]{battery SoC};

\draw[sig,black!65] (plant.south)--(mocap.north);

\draw[sig,black!65] (mocap.west)--(est.east) node[lbl,below,midway,text=black!65]{pos., orient.};

\end{tikzpicture}%
}

\begin{figure}[t]
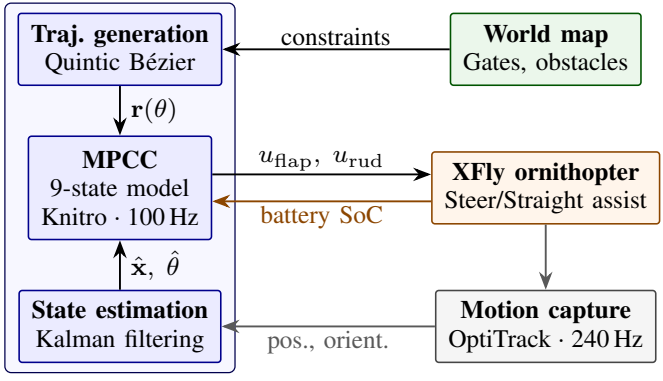

\centering
\resizebox{0.98\columnwidth}{!}{\usebox{\controlarchbox}}
\caption{Control architecture. A world map defines gate positions and obstacles that constrain the offline trajectory generator, which produces the reference trajectory~$\mathbf{r}(\theta)$ as quintic B\'{e}zier curves. Kalman filters estimate position, velocity, heading, and heading rate from OptiTrack measurements at 240\,Hz. The MPCC controller solves~\eqref{eq:mpcc} over the nine-state model at 100\,Hz, producing flapping and rudder commands. The battery state of charge adapts~$u_{\mathrm{level}}$ throughout flight. The \textcolor{blue!55!black}{blue} blocks are executed offboard on a ground station.}
\label{fig:control_arch}
\end{figure}

\subsection{Platform and Onboard Stabilization}

Experiments are conducted on the XFly ornithopter~\cite{bionicbird2024xfly}, a commercially available bird-scale flapping-wing aerial vehicle with a wingspan of approximately 38\,cm, a mass of roughly 12\,g, and maximum wingbeat frequency of $\approx$20\,Hz. The platform is actuated through two control inputs:
\begin{equation}
\mathbf{u} =
\begin{bmatrix}
u_{\text{flap}} \\
u_{\text{rud}}
\end{bmatrix},
\end{equation}
where $u_{\text{flap}} \in [0,1]$ denotes the normalized flapping intensity, which regulates wingbeat frequency, and $u_{\text{rud}} \in [-1,1]$ denotes the normalized rudder deflection, which provides heading authority.

Unlike multirotor platforms, the ornithopter does not provide independent pitch or roll control. The rudder mechanism simultaneously tensions the wings, producing a coupled roll--heading response. The onboard firmware therefore includes an IMU-based stabilization loop that regulates roll angle and roll rate towards zero, providing onboard roll stabilization while exposing flapping frequency and rudder commands to the external controller. The firmware also exposes configuration parameters controlling steering assist and straight-flight assist. In all experiments the steering assist level was set to medium and the straight-flight assist level to 50\%. Under these settings, turns at small radii induce banking and altitude loss that the onboard stabilization alone does not fully reject.

\subsection{Nine-State Ornithopter Dynamics}
\begin{figure}[t]
\centering
\includegraphics[width=0.97\columnwidth, trim={0.3cm 0.10cm 0cm 0cm}, clip]{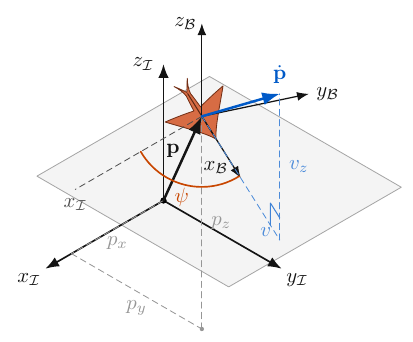}
\caption{\rev{Coordinate frames and motion variables of the bird-scale ornithopter.
The inertial frame ($x_{\mathcal I},y_{\mathcal I},z_{\mathcal I}$) and body frame
($x_{\mathcal B},y_{\mathcal B},z_{\mathcal B}$) are orthonormal and right-handed;
$\mathbf{p}=(p_x,p_y,p_z)$ is the position. The velocity $\dot{\mathbf p}$
decomposes into a horizontal component of magnitude $v$ (the airspeed), lying in
the shaded plane parallel to $x_{\mathcal I}$--$y_{\mathcal I}$ and making heading
angle $\psi$ with the $x_{\mathcal I}$ direction, and a vertical component $v_z$.}}
\label{fig:frames}
\end{figure}

Predictive control requires a dynamical model that captures the dominant aerodynamic couplings of flapping flight while remaining continuously differentiable and low-dimensional for real-time nonlinear optimization. We derive a cycle-averaged representation based on first-principles considerations of forward flight rather than modeling the detailed wing aerodynamics, which involve complex fluid-structure interactions.

\rev{The coordinate frames and the model state variables are illustrated in
Fig.~\ref{fig:frames}.} The state vector is defined as
\begin{equation}
\mathbf{x} =
[p_x, p_y, p_z, \psi, v, v_z, a_z, \dot{\psi}, \ddot{\psi}]^{\top},
\end{equation}
where $p_x,p_y,p_z$ denote position, $v$ represents horizontal airspeed, $v_z$ the vertical speed, $a_z$ the vertical acceleration, \rev{$\psi$ the heading angle
    defined as the direction of the horizontal velocity vector,
    $\psi = \operatorname{atan2}(\dot{p}_y, \dot{p}_x)$, and $\dot{\psi}$,
    $\ddot{\psi}$ its first and second time derivatives \textemdash{} the heading rate
    and the heading angular acceleration, respectively}. We represent $\psi$ as the direction \rev{angle} of the horizontal velocity vector (\rev{Fig.~\ref{fig:frames}}) rather than the body heading angle, as the velocity direction is the kinematically relevant quantity for trajectory tracking: it directly governs the evolution of position and is consistent with the horizontal dynamics below.

\subsubsection{Horizontal and Forward-Speed Dynamics}
Horizontal motion follows
\begin{align}
\dot{p}_x &= v\cos\psi, \\
\dot{p}_y &= v\sin\psi,
\end{align}
and forward speed evolves according to
\begin{equation}
\dot{v} =
k_{\text{T}} \, u_{\text{flap}}\max\!\left(0,\,1-\frac{v}{v_{\text{max}}}\right)
-
k_{\text{D}} \, v,
\label{eq:vdot}
\end{equation}
where $k_{\text{T}}$ is the thrust gain, $k_{\text{D}}$ is the drag coefficient, and $v_{\text{max}}$ is the terminal speed. \rev{This
    cycle-averaged effective form represents a propulsive force that grows with
    flapping effort but whose accelerating capacity diminishes with airspeed,
    vanishing at $v_{\text{max}}$, together with a linear aerodynamic drag.
    Although quasi-steady theory predicts thrust scaling approximately with the
    square of the wingbeat frequency, itself related to the command through the
    saturating map~\eqref{eq:flapfreq}, a first-order dependence on the flapping
    command fits the identified flight data over the operating envelope
    (Appendix~A).}
\rev{Because~(5) is only $C^{0}$ at $v=v_{\max}$, whereas the gradient-based
    solver assumes continuously differentiable dynamics, we replace the hard clamp
    with a $C^{\infty}$ surrogate: defining the smooth rectifier
    $\sigma_{\varepsilon}(x)=\tfrac{1}{2}\big(x+\sqrt{x^{2}+\varepsilon^{2}}\big)$,
    which recovers $\max(0,x)$ as $\varepsilon\!\to\!0$. Thus, the forward-speed dynamics
    become}
    \begin{equation}\label{eq:vdot_smooth}
    \rev{\dot{v} = k_{\mathrm{T}}\, u_{\mathrm{flap}}\,
    \sigma_{\varepsilon}\!\left(1-\frac{v}{v_{\max}}\right) - k_{\mathrm{D}}\, v,}
    \end{equation}
    \rev{with $\varepsilon$ chosen as $0.02$.}
 
\subsubsection{Vertical Dynamics}
Vertical motion is governed by
\begin{align}
\dot{p}_z &= v_z, \\
\dot{v}_z &= a_z - k_{\psi z}\dot{\psi}^2,
\label{eq:vzdot}
\end{align}
where the term $k_{\psi z}\dot{\psi}^2$ captures turn-induced altitude loss\rev{, $k_{\psi z}$ being the turn--altitude coupling gain}. The flapping intensity determines a target vertical velocity 
\begin{equation}
v_{z,\text{target}} = k_z(u_{\text{flap}}-u_{\text{level}}),
\end{equation}
\rev{where $k_z$ is the
    climb-rate gain and $u_{\text{level}}$ the level-flight flapping command. This linear form is a first-order approximation of the
climb-rate response about level flight ($u_{\text{flap}}\approx u_{\text{level}}$),
with $k_z$ identified from flight data.} The target velocity  is tracked through a second-order response
\begin{equation}
\dot{a}_z =
\omega_{\text{n}}^2(v_{z,\text{target}}-v_z)
-
2\zeta\omega_{\text{n}} \, a_z,
\label{eq:azdot}
\end{equation}
\rev{with natural
    frequency $\omega_n$ and damping ratio $\zeta$}.
The motor produces less torque for a given command as the battery depletes, requiring a progressively higher flapping input to maintain level flight. Accordingly, $u_{\text{level}}$ is not a fixed constant but is updated online via a linear calibration of the form
\begin{equation}
u_{\text{level}}(b) = a_{\text{batt}}\,b + c_{\text{batt}},
\label{eq:ulevel_batt}
\end{equation}
where $b$ is the battery state of charge in percent and $a_{\text{batt}},\,c_{\text{batt}}$ are identified from flight data. This adaptation ensures that the vertical dynamics model remains accurate throughout the flight envelope. \rev{The flapping frequency is an empirically fitted saturating function of
the command ($R^2 = 0.995$), rescaled by the battery state of charge $b$
through the level-flight command $u_{\mathrm{level}}$:}
\begin{equation}
  \rev{f_{\mathrm{w}}(u_{\mathrm{flap}}, b)
  = f_{\max}\left(1 - e^{-\kappa\,u_{\mathrm{flap}}}\right)
  \frac{1 - e^{-\kappa\,u_{\mathrm{level}}(100)}}
       {1 - e^{-\kappa\,u_{\mathrm{level}}(b)}}},
  \label{eq:flapfreq}
\end{equation}
\rev{with $f_{\max}$ the maximum flapping frequency and $\kappa$ the saturation rate.}

\subsubsection{Heading Dynamics}
The heading dynamics are modeled as a third-order chain to capture the inertial lag observed during turning maneuvers:
\begin{align}
\frac{d\psi}{dt}        &= \dot{\psi}, \label{eq:psidot}\\
\frac{d\dot{\psi}}{dt}  &= \ddot{\psi}, \label{eq:psiddot}\\
\frac{d\ddot{\psi}}{dt} &= \frac{1}{\tau_{\ddot{\psi}}}\bigl(\ddot{\psi}_{\text{cmd}} - \ddot{\psi}\bigr), \label{eq:psidddot}
\end{align}
with \rev{the heading-acceleration time constant
    $\tau_{\ddot{\psi}}$} and commanded heading acceleration
\begin{equation}
\ddot{\psi}_{\text{cmd}} =
k_{\text{hdg}} \, (u_{\text{rud}} + u_{\text{rud,trim}}) \, v,
\label{eq:psi_cmd}
\end{equation}
\rev{where $k_{\text{hdg}}$ is the rudder--heading gain and
    $v$ the horizontal airspeed}.
The rudder trim $u_{\text{rud,trim}}$ compensates for an inherent heading bias arising from mechanical asymmetries in the wing and rudder linkage, ensuring that a zero rudder command produces straight flight. 

\subsection{Model Predictive Contouring Control}

MPCC tracks an arc-length-parameterized trajectory $\mathbf{r}(\theta)$ rather than a time-parameterized one, treating progress along the curve as an optimization variable. This removes the need for a precomputed speed profile and allows the optimizer to negotiate tradeoffs between trajectory accuracy and forward progress in real time.

\subsubsection{Error Decomposition}

\begin{figure}
\centering
  \includegraphics[width=0.45\textwidth, trim={1.0cm -0.5cm 0 -0.5cm}, clip]{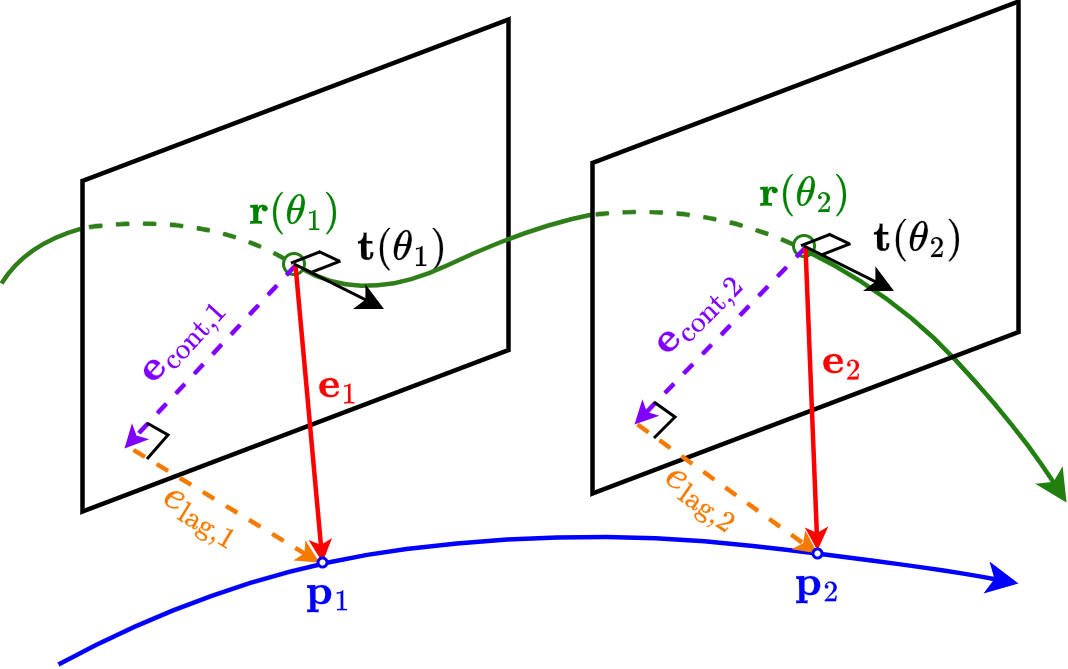}
  \caption{MPCC error decomposition at two prediction steps. The position error $\mathbf{e}_k = \mathbf{p}_k - \mathbf{r}(\theta_k)$ is projected onto the trajectory tangent $\mathbf{t}(\theta_k)$ to yield the lag (along-track) component $e_{\text{lag}}$ and the contouring (cross-track) component $\mathbf{e}_{\text{cont}}$. The optimizer penalizes contouring error to keep the vehicle near the trajectory while minimizing lag error and maximizing progress along~$\theta$.}
  \label{fig:mpcc_error}
\end{figure}

Let
\begin{equation}
\mathbf{e}_k = \mathbf{p}_k - \mathbf{r}(\theta_k)
\end{equation}
denote the position error at prediction step $k$, and let $\mathbf{t}(\theta_k)$ be the unit tangent to the trajectory. The error is decomposed into a lag (along-track) component and a contouring (cross-track) component (Fig. \ref{fig:mpcc_error}):
\begin{align}
e_{\text{lag},k} &= \mathbf{e}_k \cdot \mathbf{t}(\theta_k), \\
\mathbf{e}_{\text{cont},k} &= \mathbf{e}_k - e_{\text{lag},k}\,\mathbf{t}(\theta_k).
\end{align}

A feedforward flapping target anticipates the effort required by the local climb angle $\gamma(\theta_k)$ of the reference trajectory:
\begin{equation}
u_{\text{flap},k}^* = u_{\text{level}} + k_\gamma \, \gamma(\theta_k),
\label{eq:flap_target}
\end{equation}
\rev{where $k_\gamma$ is the climb-angle feedforward gain.}

\subsubsection{Optimization Problem}
At each control cycle the following nonlinear program is solved over the state trajectory $\mathbf{X} = \{\mathbf{x}_k\}_{k=0}^{N}$, control sequence $\mathbf{U} = \{\mathbf{u}_k\}_{k=0}^{N-1}$, progress parameter $\boldsymbol{\Theta} = \{\theta_k\}_{k=0}^{N}$, and progress speed $\mathbf{V}_\theta = \{v_{\theta,k}\}_{k=0}^{N-1}$:

\begin{subequations}\label{eq:mpcc}
\begin{align}
\min_{\mathbf{X},\, \mathbf{U},\, \boldsymbol{\Theta},\, \mathbf{V}_\theta} \;\; & \sum_{k=0}^{N-1} \Big( \;
  q_c\,\|\mathbf{e}_{\text{cont},k}\|^2
+ q_l\,e_{\text{lag},k}^2 \nonumber \\
&\quad - q_p\,v_{\theta,k}
+ q_r\,u_{\text{rud},k}^2 \nonumber \\
&\quad + q_f\,(u_{\text{flap},k} - u_{\text{flap},k}^*)^2 \; \Big)
\label{eq:mpcc_cost} \\[6pt]
\text{s.t.} \quad
& \mathbf{x}_0 = \hat{\mathbf{x}}, \quad \theta_0 = \hat{\theta},
\label{eq:mpcc_init} \\[3pt]
& \mathbf{x}_{k+1} = \mathbf{x}_k + \Delta t \, f(\mathbf{x}_k, \mathbf{u}_k),
\label{eq:mpcc_dyn} \\[3pt]
& \theta_{k+1} = \theta_k + \Delta t \, v_{\theta,k},
\label{eq:mpcc_theta} \\[3pt]
& u_{\text{flap},k} \in [0,\; 1],
\label{eq:mpcc_uflap} \\[3pt]
& u_{\text{rud},k} \in [-1,\; 1],
\label{eq:mpcc_urud} \\[3pt]
& v_k \geq v_{\min},
\label{eq:mpcc_vmin} \\[3pt]
& v_{\theta,k} \in [0,\; v_{\theta}^{\max}],
\label{eq:mpcc_vtheta}
\end{align}
\end{subequations}

\noindent for $k = 0, \dots, N\!-\!1$, where $\hat{\mathbf{x}}$ and $\hat{\theta}$ are the current state estimate and progress parameter, and $f(\cdot)$ denotes the nine-state dynamics of Section~II-B.

The cost~\eqref{eq:mpcc_cost} combines five terms. The contouring term $q_c\|\mathbf{e}_{\text{cont},k}\|^2$ penalizes cross-track deviation in all three dimensions, while $q_l\,e_{\text{lag},k}^2$ penalizes along-track error. The negative progress term $-q_p\,v_{\theta,k}$ incentivizes the vehicle to advance along the trajectory rather than slow down to reduce tracking error. The final two terms regularize control effort: rudder activity and deviation of flapping from the climb-angle-dependent feedforward target~\eqref{eq:flap_target}.

The constraints enforce the dynamics~\eqref{eq:mpcc_dyn} via Euler integration, advance the progress parameter~\eqref{eq:mpcc_theta}, and impose actuator limits on the flapping and rudder commands. The progress speed is bounded below by zero to prevent the progress parameter from regressing and above by $v_{\theta}^{\max}$.

The first control input of the optimized sequence is applied to the vehicle, and the problem is warm-started at the next cycle using the shifted previous solution. The nonlinear program is formulated in CasADi~\cite{andersson2019casadi} and solved using the Knitro solver~\cite{byrd2006knitro}, which resulted in $2\times$ lower computation time than the default IPOPT solver~\cite{wachter2006ipopt} on this problem, enabling real-time operation at 100\,Hz.

\subsubsection{Trajectory Representation}

The MPCC formulation requires a reference trajectory $\mathbf{r}(\theta)$ that is at least $C^2$ in the arc-length parameter~$\theta$. Given a sequence of three-dimensional waypoints $\{\mathbf{w}_i\}_{i=0}^{M}$, we first compute the cumulative chord-length stations $s_i = s_{i-1} + \|\mathbf{w}_i - \mathbf{w}_{i-1}\|$ and resample at uniform arc-length spacing. A cubic B-spline~\cite{deboor1978practical} is then fitted independently per coordinate axis to these uniformly spaced samples, yielding three scalar interpolants $r_x(s),\,r_y(s),\,r_z(s)$ that are each $C^2$ by construction. The unit tangent $\mathbf{t}(\theta) = \mathbf{r}'(\theta)/\|\mathbf{r}'(\theta)\|$ and climb angle $\gamma(\theta) = \arcsin(t_z)$ are obtained by symbolic differentiation of the spline basis functions.

\subsection{State Estimation}

The nine model states are reconstructed from motion-capture position and orientation measurements. A six-state constant-velocity Kalman filter~\cite{kalman1960new} on $[p_x, p_y, p_z, v_x, v_y, v_z]$ estimates position and velocity, from which horizontal airspeed is obtained as $v = \sqrt{v_x^2 + v_y^2}$.

Heading is derived from the velocity direction, $\psi = \mathrm{atan2}(v_y, v_x)$, rather than from the motion-capture quaternion. A two-state constant-heading-rate Kalman filter on $[\psi, \dot{\psi}]$ smooths this measurement. \rev{The velocity direction is not defined at stationary takeoff, so we feed
    the filter quaternion heading measurements until forward motion has stabilized
    (speed above $0.3\,$m/s, a value determined empirically).}

The remaining two states are obtained by numerical differentiation. Vertical acceleration $a_z$ is computed as the finite difference of the Kalman-filtered vertical velocity $v_z$ and smoothed with a first-order low-pass filter. Heading acceleration $\ddot{\psi}$ is similarly obtained by differentiating the filtered heading rate $\dot{\psi}$ and low-pass filtering the result.

\section{Experiments}

The proposed controller is evaluated on the XFly ornithopter across multiple trajectories designed to test the controller under varying curvature and altitude changes.

\subsection{Experimental Setup}

Experiments were conducted indoors using an OptiTrack motion capture system operating at 240\,Hz.
The MPCC controller ran at 100\,Hz with a prediction horizon of $N = 15$ steps and timestep $\Delta t = 0.1$\,s, corresponding to a 1.5\,s lookahead. \rev{The solver was executed offboard on a laptop equipped with an Intel Core i9-12900H processor, while the resulting flapping and rudder commands were transmitted to the XFly via Bluetooth. With a memory footprint of 192~MB and the observed execution time (Fig. \ref{fig:performance}), the solver currently requires a companion computer whose weight exceeds the XFly's payload capacity, making onboard deployment infeasible. Further optimization of the solver and advances in embedded computing are expected to enable fully onboard execution in future work.} All dynamical model parameters were identified from flight data collected on the XFly platform (Appendix~\ref{app:identification}) and are listed in Table~\ref{tab:model_params}. The MPCC cost weights and constraint bounds, listed in Table~\ref{tab:mpcc_params}, remained fixed across all experiments.

\begin{table}[t]
\centering
\caption{Identified dynamical model parameters.}
\label{tab:model_params}
\begin{tabular}{@{}llrl@{}}
\toprule
Symbol & Description & Value & Unit \\
\midrule
\multicolumn{4}{@{}l}{\textit{Forward-speed dynamics}} \\
$k_{\text{T}}$          & Thrust gain                       & 4.07    & m/s\textsuperscript{2} \\
$k_{\text{D}}$          & Drag coefficient                  & 0.227   & 1/s \\
$v_{\text{max}}$        & Terminal speed                    & 2.96    & m/s \\
\midrule
\multicolumn{4}{@{}l}{\textcolor{black}{\textit{Flapping frequency}}} \\
\textcolor{black}{$\kappa$} & \textcolor{black}{Flap-frequency saturation rate} & \textcolor{black}{2.64} & \textcolor{black}{--} \\
\textcolor{black}{$f_{\text{max}}$} & \textcolor{black}{Maximum wingbeat frequency} & \textcolor{black}{20} & \textcolor{black}{Hz} \\
\midrule
\multicolumn{4}{@{}l}{\textit{Vertical dynamics}} \\
$k_z$                   & Climb-rate gain                   & 1.6     & m/s \\
$k_{\psi z}$            & Turn--altitude coupling            & 0.075   & m/rad\textsuperscript{2} \\
$\omega_{\text{n}}$     & Vertical natural frequency        & 4.5     & rad/s \\
$\zeta$                 & Vertical damping ratio            & 0.25    & -- \\
$u_{\text{level}}$      & Level-flight flap (nominal)       & 0.70    & -- \\
$a_{\text{batt}}$       & Battery--flap slope               & $-$5.49$\times 10^{-3}$ & 1/\% \\
$c_{\text{batt}}$       & Battery--flap intercept           & 1.021   & -- \\
\midrule
\multicolumn{4}{@{}l}{\textit{Heading dynamics}} \\
$k_{\text{hdg}}$        & Rudder--heading gain              & $-$17.0 & rad/(s$\cdot$m) \\
$\tau_{\ddot{\psi}}$    & Heading-acceleration time constant & 0.15    & s \\
$u_{\text{rud,trim}}$   & Rudder trim                       & 0.075   & -- \\
\bottomrule
\end{tabular}
\end{table}

\begin{table}[t]
\centering
\caption{MPCC controller parameters.}
\label{tab:mpcc_params}
\begin{tabular}{@{}llr@{}}
\toprule
Symbol & Description & Value \\
\midrule
\multicolumn{3}{@{}l}{\textit{Cost weights}} \\
$q_c$                     & Contouring (cross-track)       & 250 \\
$q_l$                     & Lag (along-track)              & 50 \\
$q_p$                     & Progress                       & 0.1 \\
$q_r$                     & Rudder effort                  & 50 \\
$q_f$                     & Flap feedforward deviation     & 30 \\
\midrule
\multicolumn{3}{@{}l}{\textit{Feedforward and horizon}} \\
$k_\gamma$                & Climb-angle feedforward gain   & 2.0 \\
$N$                       & Prediction horizon             & 15 \\
$\Delta t$                & Prediction timestep            & 0.1\,s \\
\midrule
\multicolumn{3}{@{}l}{\textit{Bounds}} \\
$v_{\min}$                & Min.\ airspeed                 & 0.0\,m/s \\
$v_{\theta}^{\max}$       & Max.\ progress speed           & 5.0\,m/s \\
\bottomrule
\end{tabular}
\end{table}

\begin{figure}[t]
\centering
\includegraphics[width=\columnwidth, trim=0 0 -0.5cm 0, clip]{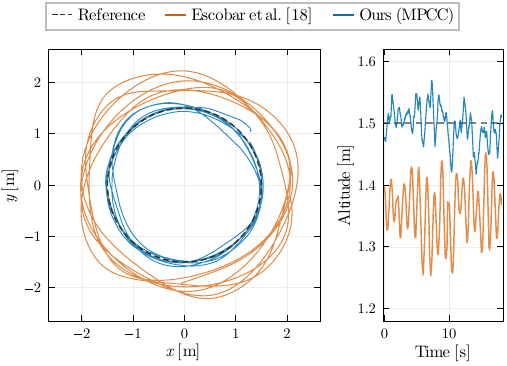}
\caption{Circle tracking comparison with Escobar et al.~\cite{escobar2025dynamical} on the XFly ornithopter ($r=1.5\,$m, $z_d=1.5\,$m). \textit{Left:} XY trajectory. \textit{Right:} altitude tracking.}
\label{fig:circle_comparison}
\end{figure}
\begin{table}[!h]
\centering
\caption{Circle tracking error comparison ($r=1.5\,$m, $z_d=1.5\,$m).}
\label{tab:circle_comparison}
\begin{tabular}{@{}l cc cc cc@{}}
\toprule
 & \multicolumn{2}{c}{\textbf{XY [cm]}} & \multicolumn{2}{c}{\textbf{Alt.\ [cm]}} & \multicolumn{2}{c}{\textbf{3D [cm]}} \\
\cmidrule(lr){2-3} \cmidrule(lr){4-5} \cmidrule(lr){6-7}
 & $\mu\pm\sigma$ & max & $\mu\pm\sigma$ & max & $\mu\pm\sigma$ & max \\
\midrule
Ours       & $5.7\pm4.8$ & $22$ & $2.3\pm1.8$ & $8.2$ & $6.5\pm4.7$ & $23$ \\
\cite{escobar2025dynamical} & $53\pm14$ & $85$ & $14\pm4.6$ & $25$ & $55\pm14$ & $89$ \\
\bottomrule
\end{tabular}
\end{table}

\subsection{Circular Trajectory}

The first experiment evaluates tracking performance on a circular trajectory of radius 1.5\,m and of constant altitude (1.5\,m). 
The controller achieved a mean cross-track error of 6.5\,cm while maintaining airspeeds between 2.0 and 3\,m/s, an 8.5$\times$ improvement over the previous state of the art on the platform \cite{escobar2025dynamical} (Fig. \ref{fig:circle_comparison}, Tab. \ref{tab:circle_comparison}). The results reported for \cite{escobar2025dynamical} are taken from the original publication, which used the same XFly platform with the same onboard stabilization firmware. Both setups use OptiTrack motion capture with sub-centimeter accuracy, ensuring that the reported improvement reflects the outer-loop control strategy rather than sensing or platform differences.

\subsection{Racing Trajectories}

\begin{figure*}[t]
\centering
\begin{minipage}{0.49\textwidth}
\centering
\setlength{\fboxsep}{0pt}%
\setlength{\fboxrule}{0.4pt}%
\fbox{%
\begin{minipage}{\dimexpr\linewidth-2\fboxrule\relax}%
\includegraphics[width=\linewidth, trim={0cm 0cm 0cm 0cm}, clip]{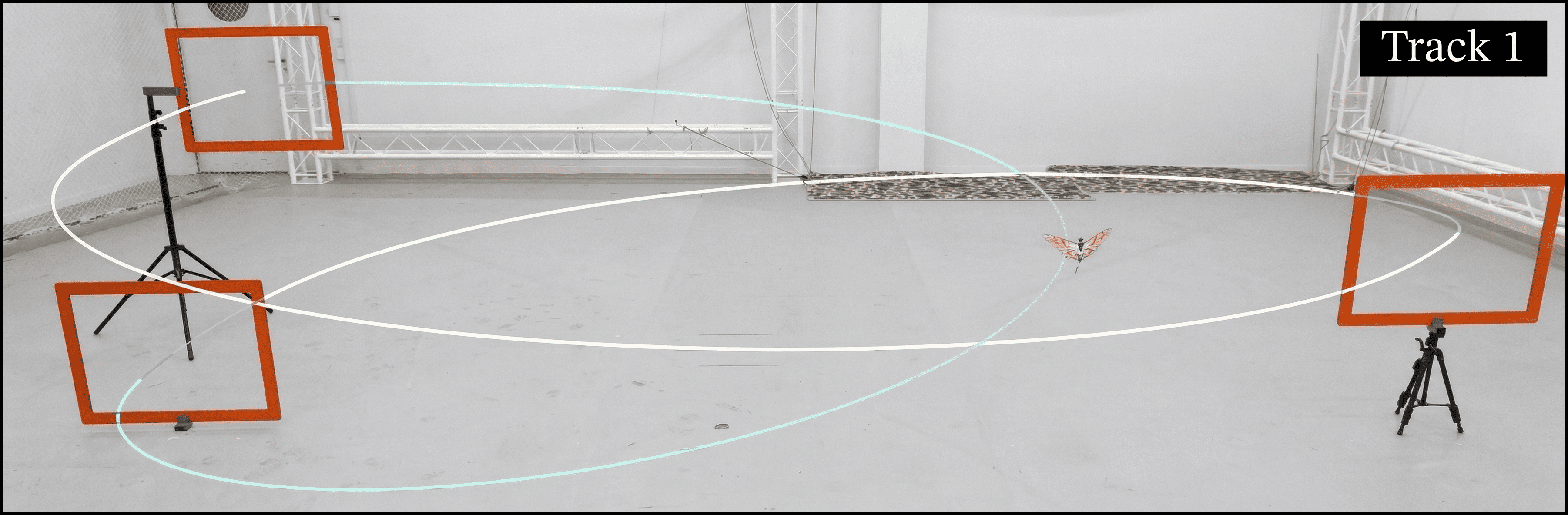}%
\par\nointerlineskip
{\color{black}\rule{\linewidth}{0.4pt}}%
\par\nointerlineskip
\includegraphics[width=\linewidth, trim={-0.25cm -0.2cm -0.5cm -0.4cm}, clip]{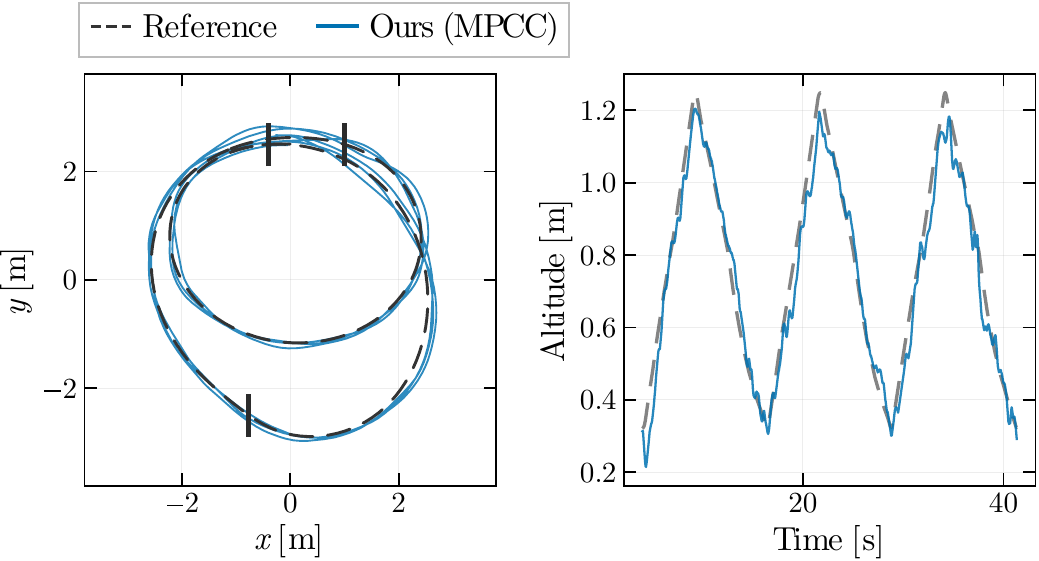}%
\end{minipage}%
}%
\end{minipage}%
\hfill
\begin{minipage}{0.49\textwidth}
\centering
\setlength{\fboxsep}{0pt}%
\setlength{\fboxrule}{0.4pt}%
\fbox{%
\begin{minipage}{\dimexpr\linewidth-2\fboxrule\relax}%
\includegraphics[width=\linewidth, trim={0cm 0cm 0cm 0cm}, clip]{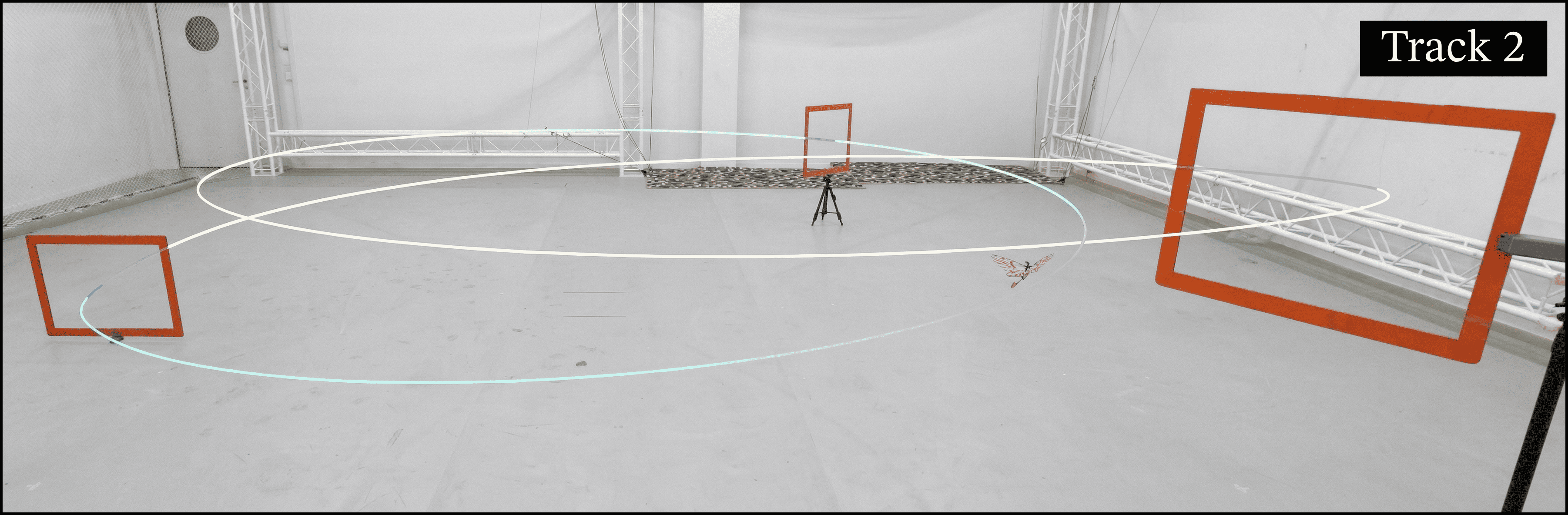}%
\par\nointerlineskip
{\color{black}\rule{\linewidth}{0.4pt}}%
\par\nointerlineskip
\includegraphics[width=\linewidth, trim={-0.25cm -0.2cm -0.5cm -0.4cm}, clip]{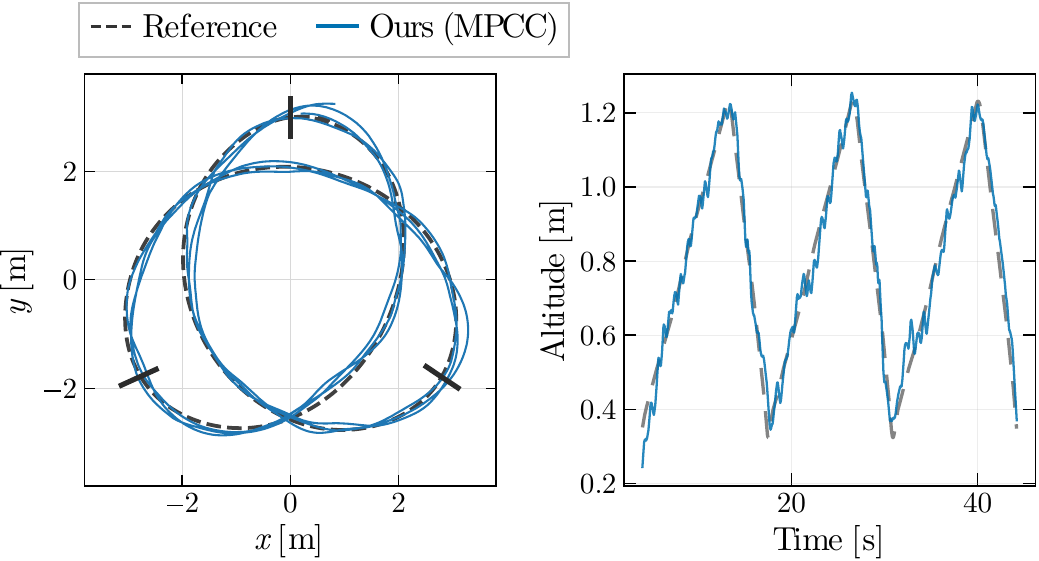}%
\end{minipage}%
}%
\end{minipage}

\caption{Experimental flight results on two three-dimensional racing trajectories over 3 laps. \textit{Top:} time-lapse photographs of the ornithopter tracking the reference trajectory through gate obstacles. \textit{Bottom:} top-down XY view (left) and altitude over time (right) for each track, comparing the MPCC-tracked trajectory (solid blue) against the reference trajectory (dashed black). Gate positions are indicated by black markers.}
\label{fig:flight_results}
\end{figure*}

Two racing trajectories consisting of three-gate closed
loops were evaluated over 3 laps (Fig. \ref{fig:flight_results}). These trajectories introduce
altitude variations between 0.3\,m and 1.25\,m, and thus the turning radius was limited to 1.8\,m to ensure feasibility and adhere to the physical limits of the platform. The
reference trajectories were generated offline by optimizing smooth quintic
B\'{e}zier curves through the gate sequence; full details are provided
in Appendix~\ref{app:traj_opt}.

\begin{figure}[t]
\centering
\includegraphics[width=\columnwidth, trim={0cm 0cm -0.5cm 0cm}, clip]{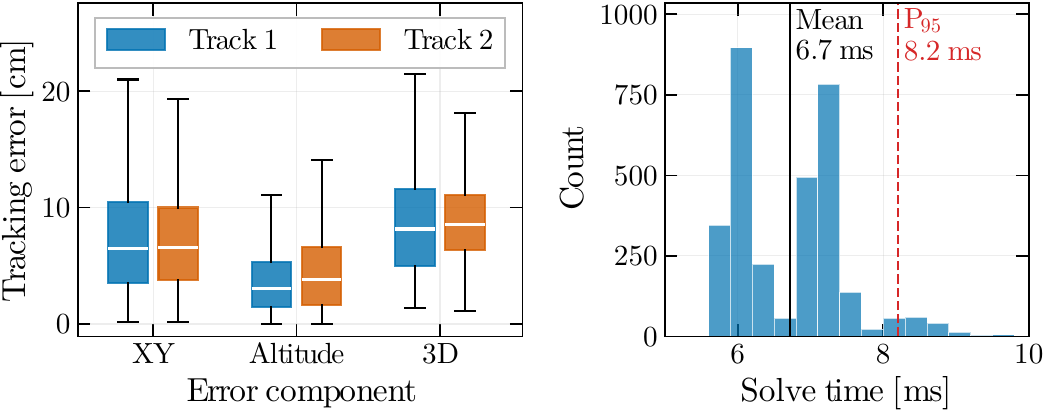}
\caption{Quantitative performance analysis over 3 laps on 2 tracks. Left: box plots of XY, altitude, and 3D tracking errors for both tracks, with median errors below 8\,cm in XY and 4\,cm in altitude. Right: distribution of MPC solve times for Track 1 (KNITRO solver) with a mean of 6.7\,ms and 95th percentile of 8.2\,ms, comfortably within the 10\,ms control period.}
\label{fig:performance}
\end{figure}

The mean contouring error was 8.5\,cm for track 1 and 8.8\,cm for track 2, while computation time never exceeded the control time step of 10\,ms (Fig. \ref{fig:performance}).

\subsection{Ablation Study}

\rev{We substituted our model with the simplified model of~\cite{escobar2025dynamical}, to isolate the effect of the model from the
control method. The resulting nonlinear program solves in roughly 40 ms per cycle (25 Hz), and the
vehicle could not be stabilized in flight despite extensive re-tuning, whereas our model runs at 100 Hz;
this configuration is therefore omitted from the comparisons below, which include only controllers
that achieved stable flight on the platform.}
     Four ablated configurations are evaluated on \rev{the circular trajectory} to isolate the contribution of each design choice. Each modifies exactly one component; all other settings remain identical. (i) Replacing the speed-dependent heading command~\eqref{eq:psi_cmd} with a fixed nominal airspeed $\bar{v}$, (ii) removing the turn-induced altitude coupling $k_{\psi z}\dot{\psi}^2$ from~\eqref{eq:vzdot}, (iii) collapsing the vertical dynamics to second order by removing the $a_z$ state \rev{from~(\ref{eq:vzdot}) and~(\ref{eq:azdot}), replacing them with the first-order
    climb-rate response
    $\dot{v}_z = \omega_z\,(v_{z,\text{target}} - v_z) - k_{\psi z}\dot{\psi}^{2}$ with $\omega_z$ the first-order climb-rate bandwidth
    }, and (iv) collapsing the third-order heading chain to second order by removing the $\ddot{\psi}$ state \rev{from~\eqref{eq:psiddot} and~\eqref{eq:psidddot}} so that the rudder maps directly to $\dot{\psi}$. \rev{The systematic order reduction in (iii) and (iv) tests whether the vertical and heading dynamics are each modeled at the minimal order required for stable control, i.e., whether the model orders are necessary rather than arbitrary.}

Ablations (iii) and (iv), which reduce the vertical and heading dynamics to second order respectively, produced large control oscillations that led to flight instability and crashes. Without the $a_z$ state the optimizer commands aggressive flapping inputs that the vertical inertia cannot follow, destabilizing altitude. Similarly, removing the $\ddot{\psi}$ lag state causes abrupt rudder reversals that trigger divergent heading oscillations. These two configurations are therefore excluded from the quantitative comparison. Ablations (i) and~(ii) were evaluated over 40~laps on the circular trajectory; the results are summarized in Table~\ref{tab:ablation}. Both ablations significantly degrade tracking accuracy ($p < 10^{-3}$, Welch's $t$-test), confirming that the speed-dependent turning model and turn-altitude coupling are each essential to the controller's performance.

\begin{table}[h]
\centering
\caption{Ablation study: mean cross-track error (cm) on the circular trajectory over 40 laps. $p$-values are from a Welch's $t$-test against the full model.}
\label{tab:ablation}
\begin{tabular}{@{}lccc@{}}
\toprule
Configuration & Mean [cm] & Std [cm] & $p$-value \\
\midrule
Full model                      & 7.0 & 1.8 & --- \\
(i) Speed-indep.\ turning      & 10.1 & 4.1 & \rev{$5.6\!\times\!10^{\expminus 5}$} \\
(ii) No turn-alt.\ coupling    & 12.3 & 7.9 & \rev{$1.6\!\times\!10^{\expminus 4}$} \\
\rev{(iii) 2nd-order vertical (no $a_z$)}      & \multicolumn{3}{c}{\rev{unstable (crash)}} \\
\rev{(iv) 2nd-order heading (no $\ddot{\psi}$)} & \multicolumn{3}{c}{\rev{unstable (crash)}} \\
\bottomrule
\end{tabular}
\end{table}

\section{Conclusion}

\rev{This letter} presented a Model Predictive Contouring Control (MPCC) framework for bird-scale ornithopters that combines a compact nine-state dynamical model with a trajectory-parameterized optimization that eliminates the need for precomputed speed profiles while managing the aerodynamic couplings that cascaded controllers neglect. 
Experimental validation on the XFly ornithopter demonstrated a mean cross-track error of 6.5\,cm on circular trajectories (an 8.5$\times$ improvement over the prior state of the art~\cite{escobar2025dynamical}) and below 9\,cm on three-dimensional racing trajectories with altitude variations, all at speeds up to 3\,m/s with solve times comfortably within the 10\,ms control period. Furthermore, we conducted an ablation study to quantify the contribution of each modeling choice to tracking accuracy. 
A natural next step is to extend the framework to online motion planning for single and multiple flapping-wing vehicles, by adapting the quadrotor planners in \cite{toumieh2021high, toumieh2022decentralized, toumieh2024high} to the presented model.

More broadly, these results establish the trajectory-tracking precision that higher-level autonomy — obstacle avoidance, multi-vehicle coordination, perception-guided flight — requires, bringing bird-scale flapping-wing drones within reach of the autonomous behaviors long demonstrated on rotary-wing platforms.
\appendices

\section{Parameter Identification}
\label{app:identification}

The model parameters (Table~\ref{tab:model_params}) are identified sequentially from dedicated flight maneuvers, each stage using previously identified parameters. First, a continuous level flight from full charge to depletion identifies $a_{\text{batt}}$ and $c_{\text{batt}}$ by linear regression of the flapping command required to maintain $v_z \approx 0$ against battery state of charge, establishing the time-varying baseline $u_{\text{level}}(b)$ used in all subsequent stages. Straight-flight segments at varying flapping levels then identify $k_{\text{T}}$, $k_{\text{D}}$, and $v_{\text{max}}$ via least-squares on~\eqref{eq:vdot_smooth}, including transient phases. The rudder trim $u_{\text{rud,trim}}$ is extracted from straight segments as the command producing zero mean heading rate. Rudder step responses then identify $k_{\text{hdg}}$ and $\tau_{\ddot{\psi}}$ by fitting~\eqref{eq:psidddot}--\eqref{eq:psi_cmd}, using Kalman-filtered airspeed in~\eqref{eq:psi_cmd}. Flapping steps during straight flight ($\dot{\psi} \approx 0$) identify $k_z$, $\omega_{\text{n}}$, $\zeta$ from the vertical response~\eqref{eq:vzdot}--\eqref{eq:azdot}, with the concurrent speed transient predicted by the forward-speed model. Turning maneuvers isolate $k_{\psi z}$ from residual altitude loss. All parameters were jointly refined through iterative flight testing to optimize closed-loop performance.

\rev{We validate the identified model by
    its open-loop one-step prediction error on flight data: advancing the model
    one control step ($\Delta t = 0.1$\,s) from each measured state under the
    applied commands, the root-mean-square errors over all flights are 0.04\,m
    in 3D position and $3.1^{\circ}$ in heading. The rate states ($v$, $v_z$,
    $a_z$, $\dot{\psi}$, $\ddot{\psi}$) carry flapping-frequency oscillation
    that the cycle-averaged model abstracts by design; model accuracy is
    therefore quantified on position and heading, whose reported errors include
    any downstream effect of the rate states. Over the full 1.5\,s prediction
    horizon, the
    RMSE of the controller's predictions against the realized trajectory is, at
    the horizon end, 0.35\,m in 3D position and $17^{\circ}$ in heading
    (${\approx}3.8$\,m traveled).}    \section{Racing Trajectory Generation}
\label{app:traj_opt}

The racing trajectories are generated offline by optimizing smooth,
dynamically feasible closed-loop trajectories through a sequence of $n$ gates.
Each gate $G_i$ is defined by a position $\mathbf{g}_i \in \mathbb{R}^3$ and a
unit normal $\hat{\mathbf{n}}_i$ specifying the traversal direction.

\subsection{Quintic B\'{e}zier Parameterization}
Each segment connecting gate $G_i$ to $G_{i+1\!\!\mod n}$ is a quintic
B\'{e}zier curve~\cite{farin2002curves}
\begin{equation}
\mathbf{C}_i(t) = \sum_{j=0}^{5} \binom{5}{j}(1-t)^{5-j}\,t^{\,j}\,\mathbf{P}_j^{(i)},
\quad t \in [0,1],
\label{eq:bezier}
\end{equation}
with endpoints $\mathbf{P}_0^{(i)} = \mathbf{g}_i$ and $\mathbf{P}_5^{(i)} = \mathbf{g}_{i+1}$ fixed at
the gate positions, leaving the four interior control points
$\mathbf{P}_1^{(i)},\dots,\mathbf{P}_4^{(i)} \in \mathbb{R}^3$ as free variables. The
full decision vector is the concatenation of all interior control
points, giving $12\,n$ scalar unknowns.

\subsection{Optimization Problem}
The trajectory is obtained by solving
\begin{equation}
\min_{\{\mathbf{P}_j^{(i)}\}} \;\; \sum_{i=1}^{n} \int_0^1 \kappa_i(t)^2
\, \|\mathbf{C}_i'(t)\| \, dt
\;+\; \lambda\,\Phi_{\text{pen}},
\label{eq:traj_opt}
\end{equation}
where $\kappa_i(t) = \|\mathbf{C}_i' \times \mathbf{C}_i''\| / \|\mathbf{C}_i'\|^3$ is the
curvature. The penalty term $\Phi_{\text{pen}}$ enforces $C^1$ and $C^2$ continuity at segment junctions by penalizing $\|\mathbf{C}_i'(1) - \mathbf{C}_{i+1}'(0)\|^2$ and $\|\mathbf{C}_i''(1) - \mathbf{C}_{i+1}''(0)\|^2$, together with gate-alignment terms $\bigl(1 - \hat{\mathbf{n}}_i \cdot \mathbf{C}_i'(0)/\|\mathbf{C}_i'(0)\|\bigr)^2$ that encourage traversal along each gate normal. Additional penalties enforce a minimum turning radius ($1/\kappa \geq 1.8\,$m), a maximum climb angle ($|\dot{z}/\|\mathbf{C}'\|| \leq \tan 20^\circ$) and bounding-box constraints matching the flight volume.

\subsection{Initialization and Solution Procedure}

The optimizer is seeded with a heuristic initial guess that places the interior control points using the gate normals as departure and arrival directions. For each segment from gate $G_i$ (position $\mathbf{g}_i$, normal $\hat{\mathbf{n}}_i$) to gate $G_{i+1}$ (position $\mathbf{g}_{i+1}$, normal $\hat{\mathbf{n}}_{i+1}$), a characteristic distance $d = \max(0.3\,\|\mathbf{g}_{i+1} - \mathbf{g}_i\|,\; 1.0\,\text{m})$ is computed. The four interior control points are then placed as
\begin{align}
\mathbf{P}_1 &= \mathbf{g}_i + d\,\hat{\mathbf{n}}_i, \nonumber\\
\mathbf{P}_2 &= \mathbf{g}_i + 1.8\,d\,\hat{\mathbf{n}}_i + 0.3\,(\mathbf{g}_{i+1} - \mathbf{g}_i), \nonumber\\
\mathbf{P}_3 &= \mathbf{g}_{i+1} - 1.8\,d\,\hat{\mathbf{n}}_{i+1} + 0.3\,(\mathbf{g}_i - \mathbf{g}_{i+1}), \nonumber\\
\mathbf{P}_4 &= \mathbf{g}_{i+1} - d\,\hat{\mathbf{n}}_{i+1}.
\label{eq:init_cp}
\end{align}
The first and last points ($\mathbf{P}_1$, $\mathbf{P}_4$) lie along the respective gate normals and control the departure and arrival tangent directions, while the middle points ($\mathbf{P}_2$, $\mathbf{P}_3$) are offset further along the normals with an additional 30\% bias toward the opposite gate, producing a smooth S-shaped horizontal profile. The $z$-coordinates are linearly interpolated between gate heights using $z_k = (1 - \alpha_k)\,z_i + \alpha_k\,z_{i+1}$ with $\alpha_k \in \{0.2, 0.4, 0.6, 0.8\}$, and all coordinates are clamped to the flight volume bounds.

The cost landscape is multimodal---different control-point arrangements produce distinct trajectory topologies through the gates---so the problem is solved in two stages. First, differential evolution (DE)~\cite{storn1997differential}, a gradient-free population-based optimizer, explores the space globally. DE maintains a population of $N_p = 15 \times 12\,n$ candidate solutions initialized by uniform sampling within the box bounds, seeded with the heuristic guess above. At each generation, a trial vector $\mathbf{v}_i$ is constructed for every candidate $\mathbf{x}_i$ via mutation
\begin{equation}
\mathbf{v}_i = \mathbf{x}_a + F\,(\mathbf{x}_b - \mathbf{x}_c),
\label{eq:de_mutation}
\end{equation}
where $\mathbf{x}_a, \mathbf{x}_b, \mathbf{x}_c$ are distinct randomly selected candidates and the scale factor $F$ is drawn uniformly from $[0.5, 1.5]$ each generation. A trial vector is then assembled coordinate-by-coordinate via binomial crossover with probability $CR = 0.8$, and greedy selection retains whichever of the trial or parent has lower cost. The algorithm runs for 60 generations, after which L-BFGS-B~\cite{zhu1997lbfgsb} refines the best candidate using approximate second-order information for up to 800 iterations. The resulting dense waypoint sequence is converted into the $C^2$ B-spline trajectory representation of Section~II-C.

\bibliographystyle{IEEEtran}
\bibliography{IEEEabrv,bibliography}

\end{document}